\documentclass[conference]{IEEEtran}

\usepackage{cite}
\usepackage{amsmath,amssymb,amsfonts}
\usepackage{url} 
\usepackage{algorithmic}
\usepackage{booktabs}
\usepackage{multirow}
\usepackage{makecell}
\usepackage{graphicx}
\usepackage{textcomp}
\usepackage{hyperref}
\usepackage[dvipsnames]{xcolor}
\def\BibTeX{{\rm B\kern-.05em{\sc i\kern-.025em b}\kern-.08em
    T\kern-.1667em\lower.7ex\hbox{E}\kern-.125emX}}

\makeatletter
\def\thm@headfont{\bfseries}
\def\@begintheorem#1#2{\par\noindent\textbf{#1\ #2.}\ \it}
\def\@opargbegintheorem#1#2#3{\par\noindent\textbf{#1\ #2.\ (#3).}\ \it}
\makeatother

\newtheorem{rem}{\bf{Remark}}

\begin{document}

\title{TensorLLM: Tensorising Multi-Head Attention for Enhanced Reasoning and Compression in LLMs}

\author{
Yuxuan Gu, Wuyang Zhou, Giorgos Iacovides, Danilo Mandic \\
\textit{Department of Electrical and Electronic Engineering}\\
\textit{Imperial College London, United Kingdom}\\
\{yuxuan.gu21, wuyang.zhou19, giorgos.iacovides20, d.mandic\}@imperial.ac.uk
}

\maketitle

\begin{abstract}
The reasoning abilities of Large Language Models (LLMs) can be improved by structurally denoising their weights, yet existing techniques primarily focus on denoising the feed-forward network (FFN) of the transformer block, and can not efficiently utilise the Multi-head Attention (MHA) block, which is the core of transformer architectures. To address this issue, we propose a novel intuitive framework that, at its very core, performs MHA compression through a multi-head tensorisation process and the Tucker decomposition. This enables both higher-dimensional structured denoising and compression of the MHA weights, by enforcing a shared higher-dimensional subspace across the weights of the multiple attention heads. We demonstrate that this approach consistently enhances the reasoning capabilities of LLMs across multiple benchmark datasets, and for both encoder-only and decoder-only architectures, while achieving compression rates of up to $\sim 250$ times in the MHA weights, all without requiring any additional data, training, or fine-tuning. Furthermore, we show that the proposed method can be seamlessly combined with existing FFN-only-based denoising techniques to achieve further improvements in LLM reasoning performance. \footnote{The code is available at \href{https://github.com/guyuxuan9/TensorLLM}{https://github.com/guyuxuan9/TensorLLM.}}
\end{abstract}


\section{Introduction}
Large Language Models (LLM) based on the transformer architecture, such as those belonging to the GPT-series \cite{ brown2020language, achiam2023gpt} and LLaMA-series \cite{touvron2023LLaMA, llama2, dubey2024LLaMA}, have demonstrated enormous success across diverse applications in natural language processing (NLP) \cite{wang-etal-2019-learning-deep, irie19b_interspeech, liu2019text}. This is attributed to the exceedingly large size of these models and the vast amount of data for their training. Indeed, transformer models with more parameters or larger training datasets tend to comprehensively outperform their smaller-scale predecessors \cite{brown2020language,touvron2023LLaMA}, owing to their ability to capture more complex patterns. \par 
The success of LLMs has demonstrated that massive over-parameterization is beneficial during training \cite{overparam}, however, a growing body of research suggests that transformer-based models, and neural networks (NN) in general, do not require all learnt parameters to maintain their performance \cite{sajjad2023effect, Fan2020Reducing}. This has led to the exploration of post-training compression techniques, which aim to make LLMs more efficient and practical for real-world applications, particularly in resource-constrained environments. This should be achieved without largely compromising their inference performance and generative capabilities \cite{sharma2023truth,eff_inference_3, calvi2019compression, xu2023tensorgpt}. \par 
Given that LLMs are grossly overparameterised, it comes as no surprise that the reasoning abilities of LLMs can be maintained, or even improved, when structurally compressing their parameters. For instance, in the LAyer-SElective Rank reduction (LASER) model \cite{sharma2023truth}, the authors demonstrated that by applying singular value decomposition (SVD) to the individual matrices, and subsequently removing the factors corresponding to the smallest singular values, one can improve the reasoning performance of LLMs. This is particularly the case when compressing the feed-forward network (FFN) weight matrices. The authors argued that by structurally compressing the individual weight matrices, they were essentially removing the weight noise caused by the randomness introduced during training. However, LASER applies SVD to only individual weight matrices, and is unable to exploit the information shared between the weight matrices. \par 
To overcome this limitation, authors in \cite{luo2024trawl} introduced the Tensor Reduced and Approximated Weights (TRAWL) model, which adopts a tensor-based approach by naively stacking weight matrices from either the multi-head attention (MHA) block or the FFN block into a higher-dimensional tensor, before applying tensor decompositions. While TRAWL can leverage the inherent tensor structure within the weights to exploit the inter-matrix correlations and can even outperform LASER in some experiments, it was found to be only effective when denoising the FFN blocks and not the MHA blocks. \par 

The success of the existing approaches for denoising FFN blocks \cite{sharma2023truth, luo2024trawl} has also highlighted the open problem of exploiting similar denoising benefits in the MHA blocks. Indeed, existing approaches \cite{sharma2023truth, luo2024trawl} usually achieve their best performance gains when denoising the weight matrices in the FFN blocks. However, when applied to the denoising of weights in MHA blocks, similar performance increases were not observed. This discrepancy is significant and somewhat surprising, as the MHA mechanism is widely regarded as the very core of the transformer architecture and LLMs in general. To this end, in this work, we show that the reason why prior works were not effective when applied to MHA is that they do not leverage domain knowledge — the multiple attention heads in each layer should operate in a coherent way rather than independently.

\begin{figure*}[ht]
\centering
\includegraphics[width=\textwidth]{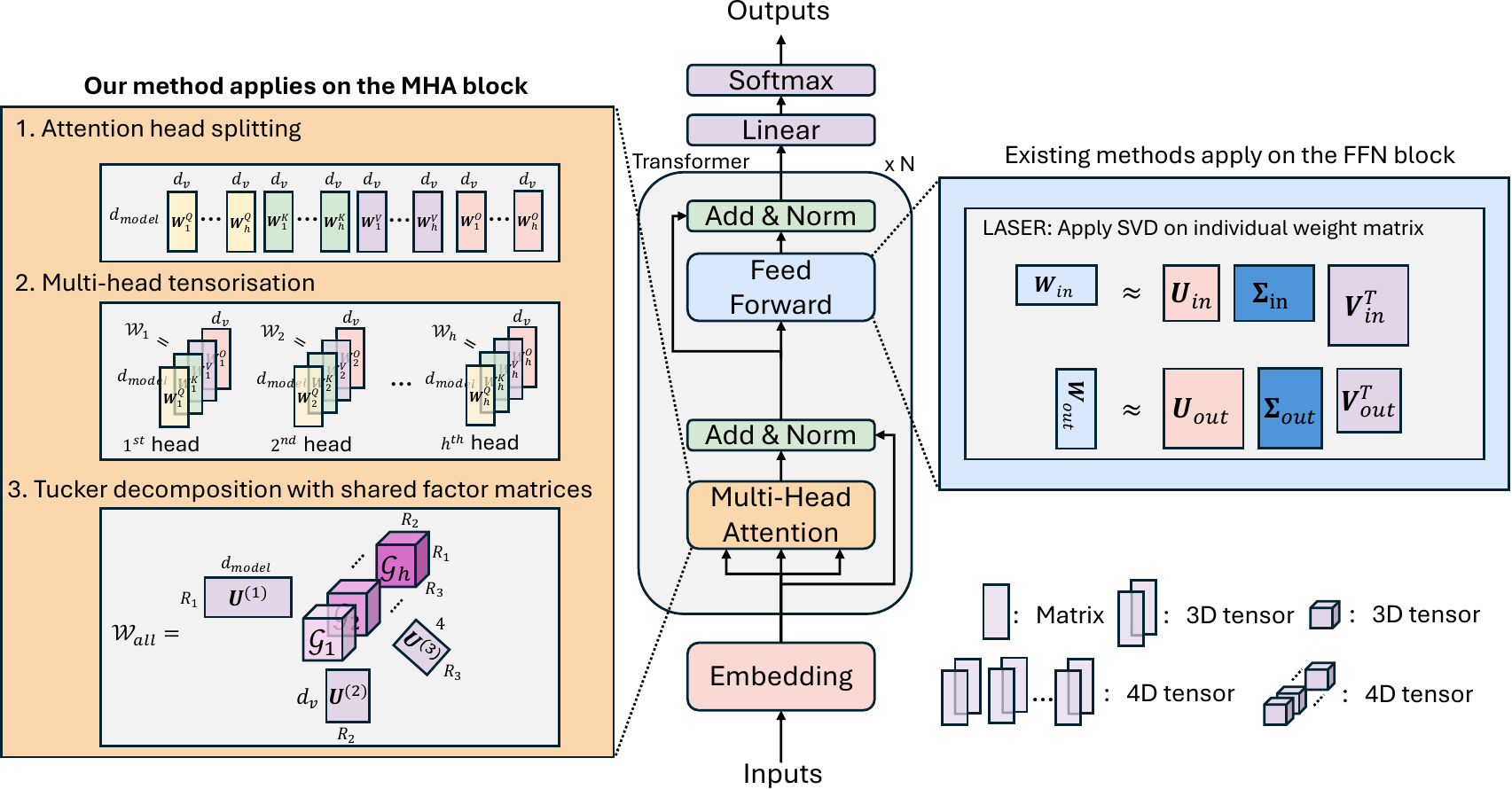} 
\caption{Structures of our proposed framework for the compression of the MHA block and the existing methods (such as LASER) which apply to the FFN block. \textbf{Left}: The three-step denoising process applied to the MHA block (\textbf{our method}): (1) split the weight matrices into multiple heads, (2) tensorise the matrices of each attention head into a 3D tensor, and (3) apply Tucker decomposition to this set of 3D tensors with a common set of factor matrices to perform denoising. \textbf{Middle}: A standard (vanilla) decoder-only or encoder-only transformer architecture. \textbf{Right}: Existing methods applied to the FFN block; an illustration of LASER.}
\label{fig:methodology}
\end{figure*}

Both design intuitions and current literature \cite{clark2019does, vig2019multiscale, vig2019analyzing} on MHA suggest that: 
\begin{enumerate}
    \item Attention heads within the same layer capture the same level of patterns;
    \item Different attention heads within the same layer learn different specialized knowledge.
\end{enumerate} More specifically, the authors in \cite{clark2019does} provided empirical evidence to support these intuitions by analyzing the Jensen-Shannon divergence between the outputs of the corresponding attention heads. Their analysis revealed distinct clusters of attention heads within transformer layers, indicating that heads in the same layer tend to focus on similar type of information, albeit with some degree of variation. Furthermore, the authors in \cite{vig2019multiscale, vig2019analyzing} visualized attention on multiple scales and demonstrated the specialised functions of different heads, such as those capturing positional and lexical patterns, detecting named entities, and identifying syntactic and semantic relations. \par

By leveraging on those intuitions and domain knowledge about MHA, we conjecture that the weights of MHA in a single transformer layer contain reasoning-related information in a subspace which is shared across multiple attention heads. We explore and elaborate this conjecture to help mitigate the limitations of existing works by answering the following question:
\begin{itemize}
    \item Can we improve the reasoning capabilities of LLMs by enforcing a shared higher-dimensional subspace among the weights of multiple attention heads within a single transformer layer?
\end{itemize} 
To answer this question, we propose a novel and intuitive framework based on the multi-head tensorisation and a special variant of the Tucker decomposition, which denoises the original MHA weights according to a shared higher-dimensional low-rank structure across multiple attention heads. By enforcing the weights of each attention head to be in a common higher-dimensional subspace characterised by a common set of Tucker factor matrices, this makes each attention head contain different information in the same subspace. In contrast to existing approaches that focus on the FFN weights only, we show that this improves the reasoning capabilities of LLMs by structurally denoising and compressing the MHA weights in a higher-dimensional format.  \par 
The main contributions of this work are threefold: 
\begin{itemize}
    \item A novel post-training weight-denoising framework is proposed based on prior knowledge, intuition, and empirical evidence about MHA, in order to improve the reasoning abilities of LLMs while simultaneously performing parameter compression. This is achieved through a unique multi-head tensorisation technique and a special variant of the Tucker decomposition applied on the MHA weight matrices. 
    \item We demonstrate that the proposed framework can be used in conjunction with existing methods that denoise the FFN layers, to achieve even greater gains in the reasoning abilities of LLMs. 
    \item Extensive experiments are conducted on four benchmark datasets across three well-established LLMs, ranging from multiple millions to billions of parameters and from encoder-only to decoder-only architectures. Our method is found to improve the reasoning abilities of both the uncompressed LLMs and existing methods that focus only on denoising the FFN layers, all without requiring any additional data, training or fine-tuning.
\end{itemize}
    
\begin{figure*}[ht]
\centering
\includegraphics[width=\textwidth]{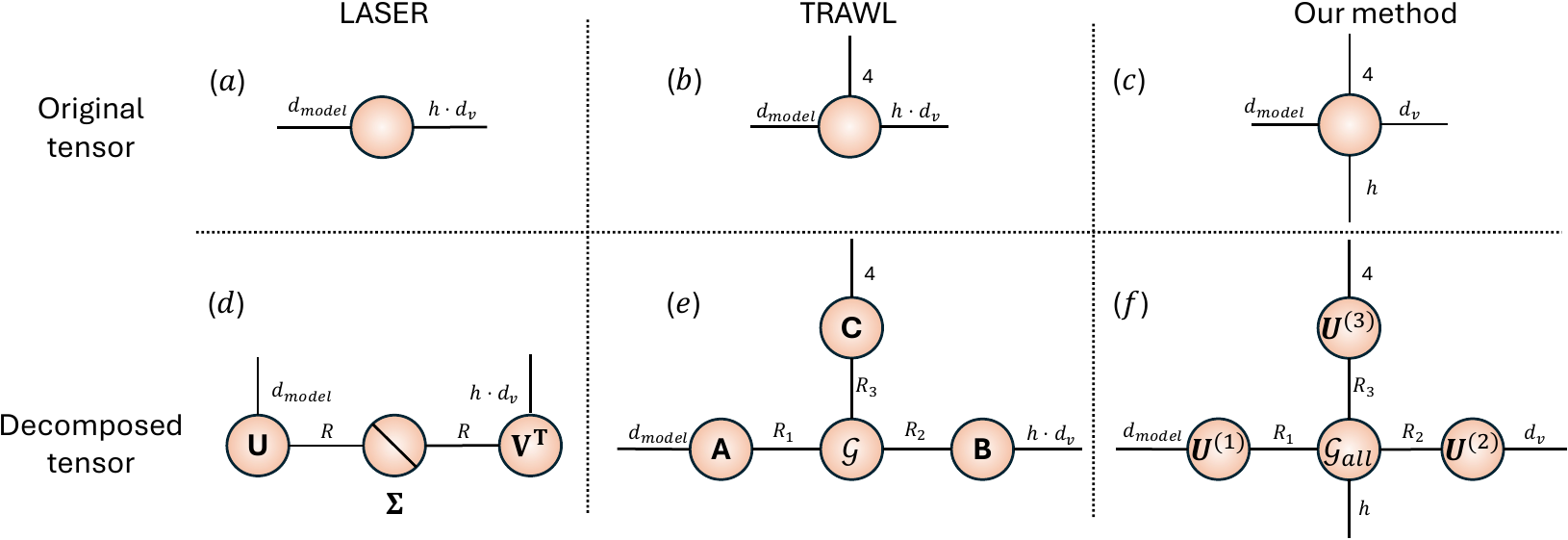} 
\caption{Tensor network \cite{orus2014practical} topologies of different decomposition methods applied to the MHA weights in a single transformer layer. Note that while both LASER and TRAWL reported performances when only applied to the FFN blocks, they could also be applied to the MHA weights. The symbol $d_{model}$ represents the embedding dimension, $h$ is the number of attention heads, while $d_v = \frac{d_{model}}{h}$ stands for the head dimension. ($\mathbf{a}, \mathbf{d}$) LASER \cite{sharma2023truth}: decomposition of a single weight matrix into $\mathbf{U} \in \mathbb{R}^{d_{model} \times R}$, $\mathbf{V} \in \mathbb{R}^{h \cdot d_v \times R}$, and a diagonal matrix $\mathbf{\Sigma} \in \mathbb{R}^{R \times R}$. ($\mathbf{b},\mathbf{e}$) TRAWL \cite{luo2024trawl}: decomposition of a 3D tensor into the factor matrices $\mathbf{A} \in \mathbb{R}^{4 \times R_3}, \mathbf{B} \in \mathbb{R}^{h \cdot d_v \times R_2}$, and $\mathbf{C} \in \mathbb{R}^{d_{model} \times R_1}$, along with a core $\mathcal{G} \in \mathbb{R}^{R_1 \times R_2 \times R_3}$. ($\mathbf{c}, \mathbf{f}$) \textbf{Our method} -- Tucker decomposition with shared factor matrices: decomposing a set of 3D tensors using Tucker decomposition while having a common set of shared factor matrices $\mathbf{U}^{(1)} \in \mathbb{R}^{d_{model} \times R_1}, \mathbf{U}^{(2)} \in \mathbb{R}^{d_{v} \times R_2} , \mathbf{U}^{(3)} \in \mathbb{R}^{4 \times R_3} $, with $\mathcal{G}_{all} \in \mathbb{R}^{R_1 \times R_2 \times R_3 \times h}$ as the core tensor.}
\label{fig:tensor_network}
\end{figure*}

\section{Background}
\subsection{Mathematical Notations}

The mathematical notations used in this paper are summarized in Table \ref{math_notation}. We adopt the same notation conventions as those in \cite{cichocki2015tensor}.

\begin{table}[htbp]
\scriptsize
  \caption{Mathematical notations}
  \label{math_notation}
  \centering
  \begin{tabular}{cc}
    \toprule
      Symbol   & Meaning \\
    \midrule
    $a$, $\mathbf{a}$, $\mathbf{A}$, $\mathcal{A}$    & Scalar, vector, matrix, tensor  \\
    $(\cdot)^T$ & Matrix Transpose \\
    $\|\cdot\|_F$ & Frobenius Norm \\
    $\mathcal{A}_{[i_1, i_2, \dots, i_N]}$ & The $(i_1, i_2, \dots, i_N)$-th element in an $N$-D tensor \\
    $\mathbf{a} \circ \mathbf{b}$ & Outer product between two vectors \\
    $\mathbf{a} \cdot \mathbf{b}$ & Inner product between two vectors \\
    $\mathcal{A} \times_n \mathbf{B}$ & Mode-$n$ product between a tensor and a matrix \\
    diag$\left(\lambda_1, \lambda_2, \dots, \lambda_R \right)$ & A diagonal matrix  \\
    diag$_N \left(\lambda_1, \lambda_2, \dots, \lambda_R \right)$ & A diagonal tensor of $N$-dimensions \\
    \bottomrule
  \end{tabular}
\end{table}

\subsection{Tensor Preliminaries}

\paragraph{Definition} A tensor $\mathcal{A} \in \mathbb{R}^{I_1 \times I_2 \times \dots \times I_N}$ is a multi-dimensional array. The number of modes (dimensions) in this tensor is its order, $N$. Its $n$-th mode has a size of $I_n$, where $n \in [1, N]$, along the $n^{th}$ dimension \cite{kolda2009tensor}. 

For example, a scalar $a$ is a $0$-dimensional tensor. By stacking multiple scalars together, we form a $1$-dimensional tensor $\mathbf{a}$, commonly referred to as a vector. Extending this further, stacking multiple vectors together results in a $2$-dimensional tensor $\mathbf{A}$, which is a matrix. Finally, stacking multiple matrices together produces a $3$-dimensional tensor $\mathcal{A}$. This process can be generalized to construct higher-dimensional tensors by iteratively stacking lower-dimensional ones.

\paragraph{Mode-$n$ product} Mode-$n$ product of a tensor $\mathcal{A} \in \mathbb{R}^{I_1 \times I_2 \times \dots \times I_N}$ and a matrix $\mathbf{B} \in \mathbb{R}^{J_n \times I_n}$ yields a tensor $\mathcal{C} \in \mathbb{R}^{I_1 \times \dots \times I_{n-1} \times J_n \times I_{n+1} \times I_N}$. This is defined mathematically as
\begin{equation}
    \mathcal{C} = \mathcal{A} \times_n \mathbf{B},
\label{eq:mode_n_product}
\end{equation}
with the element-wise definition of $\mathcal{C}$ as 
\begin{equation}
    \mathcal{C}_{[i_1,\dots,i_{n-1}, j_n, i_{n+1}, i_N]} = \sum_{i_n=1}^{I_n} \mathcal{A}_{[i_1,\dots,i_{n-1}, i_n, i_{n+1}, i_N]} \mathbf{B}_{[j_n, i_n]}.
\label{eq:mode_n_product_index_notation}
\end{equation}
\subsection{Singular Value Decomposition (SVD) }
In linear algebra, SVD factorizes any matrix $\mathbf{A} \in \mathbb{R}^{m \times n}$ into a sum of rank-1 matrices, and can be expressed as
\begin{equation}
\begin{split}
    \mathbf{A} &=  \sum_{i=1}^{r} \sigma_i \mathbf{u}_i \mathbf{v}_i^T = \sum_{i=1}^{r} \sigma_i \left( \mathbf{u}_i \circ \mathbf{v}_i \right)\\
    &= \mathbf{U} \mathbf{\Sigma} \mathbf{V}^T,
\end{split}
\label{eq:SVD}
\end{equation}
where $\mathbf{u}_i \in \mathbb{R}^m$ is an eigenvector of $\mathbf{A}\mathbf{A}^T$, and $\mathbf{U} = \left[\mathbf{u}_1, \mathbf{u}_2, \dots, \mathbf{u}_m \right] \in \mathbb{R}^{m \times m}$ is a matrix consisting of a set of orthonormal eigenvectors in $\mathbb{R}^{m}$. Similarly, $\mathbf{v}_i \in \mathbb{R}^n$ is an eigenvector of $\mathbf{A}^T\mathbf{A}$, and $\mathbf{V} = \left[\mathbf{v}_1, \mathbf{v}_2, \dots, \mathbf{v}_n \right] \in \mathbb{R}^{n \times n}$ is a matrix consisting of a set of orthonormal eigenvectors in $\mathbb{R}^{n}$. Moreover, $ \mathbf{u}_i \circ \mathbf{v}_i$ denotes the outer product between two vectors, $\sigma_i$ are the singular values of $\mathbf{A}$, which are the square root of the eigenvalues of $\mathbf{A}^T\mathbf{A}$, and $\mathbf{\Sigma} = \text{diag}\left( \sigma_1, \sigma_2, \dots, \sigma_r \right) \in \mathbb{R}^{m \times n}$ is a diagonal matrix, where $r$ is the rank of matrix $\mathbf{A}$. The SVD performs low rank approximation by discarding the rank-1 factors associated with the smallest singular values, which likely contain the noisy component of the data and less useful information, thus performing denoising.

\subsection{Tucker decomposition}

The Tucker decomposition \cite{tucker1966some} is a generalisation of SVD \cite{de2000multilinear, vasilescu2002multilinear}, which decomposes the original tensor into a product of factor matrices and a smaller core tensor, as shown in Eq.  (\ref{eq:tucker}). Thus, Tucker decomposition enables the denoising of the original tensor by approximating the original tensor using a higher-dimensional low-rank structure, and is defined as
\begin{equation}
\begin{split}
    \mathcal{T} &= \sum_{r_1=1}^{R_1} \sum_{r_2=1}^{R_2} \dots \sum_{r_N=1}^{R_N} \mathcal{G}_{[r_1, r_2, \dots, r_N]} \\
    & \quad \ \left(\mathbf{u}^{(1)}_{r_1} \circ \mathbf{u}_{r_2}^{(2)} \circ \dots \circ \mathbf{u}_{r_N}^{(N)} \right)\\
    &= \mathcal{G} \times_1 \mathbf{U}^{(1)} \times_2 \mathbf{U}^{(2)} \times_3 \dots \times_N \mathbf{U}^{(N)}.
\end{split}
\label{eq:tucker}
\end{equation}
In this formulation, $\mathcal{T} \in \mathbb{R}^{I_1 \times I_2 \times \dots \times I_N}$ is an $N$-dimensional tensor. The core tensor, $\mathcal{G}_{[r_1, r_2, \dots, r_N]} \in \mathbb{R}^{R_1 \times R_2 \times \cdots \times R_N}$, represents a scaling coefficient, and $\mathbf{u}_{r_n}^{(n)} \in \mathbb{R}^{I_n}$ denotes a factor vector for the $n$-th dimension, while the factor matrix for the $n$-th dimension is defined as $\mathbf{U}^{(n)} = \left[ \mathbf{u}_1^{(n)}, \mathbf{u}_2^{(n)}, \dots \mathbf{u}_{R_n}^{(n)} \right] \in \mathbb{R}^{I_n \times R_n}$ for $1\leq n \leq N$. The factor matrices in the Tucker decomposition characterise the projection of the original tensor onto a subspace. The $N$-tuple $(R_1, R_2, \dots, R_N)$, commonly referred to as the multilinear ranks, is a generalization of matrix rank to higher-dimensions and is usually treated as hyperparameters. Identifying the optimal set of ranks efficiently is an active area of research, with numerous recent studies focusing on advanced methods for tensor rank search \cite{iacovides2024towards, 10655696}. Tucker decomposition enables compression in terms of the parameter count by setting $R_n \ll I_n$ for $1\leq n \leq N$. 

\subsection{Transformer Architecture}
Current transformer architectures \cite{vaswani2017attention} used in LLMs can be categorized into the encoder-only and decoder-only categories and often comprise $N$ identical transformer layers, as illustrated in the middle section of Fig. \ref{fig:methodology}. Each layer consists of two primary components: an MHA block and an FFN block. To enhance stability and facilitate learning, layer normalization \cite{lei2016layer} and residual connections are also applied between these blocks.

\paragraph{Multi-Head Attention (MHA)}  

Single-head learnable self-attention is computed with the input query ($\mathbf{Q} \in \mathbb{R}^{L \times d_{model}}$), key ($\mathbf{K} \in \mathbb{R}^{L \times d_{model}}$), and value ($\mathbf{V} \in \mathbb{R}^{L \times d_{model}}$) matrices, defined as \cite{vaswani2017attention}:
\begin{multline} 
    \text{Attention}(\mathbf{Q}\mathbf{W}^Q_i, \mathbf{K}\mathbf{W}^K_i, \mathbf{V}\mathbf{W}^V_i) = \\ 
\text{softmax}\left(\frac{\left(\mathbf{Q}\mathbf{W}^Q_i \right) \left(\mathbf{K}\mathbf{W}^K_i \right)^T}{\sqrt{d_{v}}}\right) \left(\mathbf{V}\mathbf{W}^V_i \right),
\label{eq:self_attention} 
\end{multline}
where $\mathbf{W}^Q_i \in \mathbb{R}^{d_{model} \times d_{v}}, \mathbf{W}^K_i \in \mathbb{R}^{d_{model} \times d_{v}}$ and $\mathbf{W}^V_i \in \mathbb{R}^{d_{model} \times d_{v}}$, with $1\leq i \leq h$, are learnable projection matrices for a single attention head. Here, $L$ represents the sequence length, $d_{model}$ denotes the dimensionality of the embeddings of the model, $h$ is the number of heads, and $d_v = \frac{d_{model}}{h}$ designates the dimensionality of each head.

The primary role of multi-head attention \cite{vaswani2017attention} is to enhance the  ability of the model to capture complex patterns in the latent space, whereby each head independently learns a distinct representation of the input data. In the case of a $h$-head attention block, the outputs from all attention heads are concatenated and projected back onto the original latent space by an output projection matrix $\mathbf{W}^O$, as
\begin{equation}
\begin{split}
   & \text{MultiHead}(\mathbf{Q}, \mathbf{K}, \mathbf{V}) = \text{Concat}(\text{head}_1, \ldots, \text{head}_h) \mathbf{W}^O, \\
 & \text{where} \  \text{head}_i = \text{Attention}(\mathbf{Q}\mathbf{W}^Q_i, \mathbf{K}\mathbf{W}^K_i, \mathbf{V}\mathbf{W}^V_i).
\end{split}
\label{eq:MHA}
\end{equation}
Therefore, the four weight matrices in the Attention block are respectively the query projection weights $\mathbf{W}^Q = \left[\mathbf{W}^Q_1, \mathbf{W}^Q_2, \ldots, \mathbf{W}^Q_h \right] \in \mathbb{R}^{d_{model} \times h \cdot d_v}$, the key projection weights $\mathbf{W}^K = \left[\mathbf{W}^K_1, \mathbf{W}^K_2, \ldots, \mathbf{W}^K_h \right] \in \mathbb{R}^{d_{model} \times h \cdot d_v}$, the value projection weights $\mathbf{W}^V = \left[\mathbf{W}^V_1, \mathbf{W}^V_2, \ldots, \mathbf{W}^V_h \right] \in \mathbb{R}^{d_{model} \times h \cdot d_v}$, and the output projection weights $\mathbf{W}^{O} = \left[\mathbf{W}^O_1, \mathbf{W}^O_2, \ldots, \mathbf{W}^O_h \right] \in \mathbb{R}^{h \cdot d_v \times d_{model}}$ . We refer to $\mathbf{W}^Q, \mathbf{W}^K, \mathbf{W}^V, \text{and } \mathbf{W}^O$ as the MHA weight matrices. \par 
By representing the MHA weights as tensor network diagrams \cite{orus2014practical}, Fig. \ref{fig:tensor_network} illustrates the difference between LASER, TRAWL and the Tucker decomposition with shared factor matrices used in our proposed method (See section \ref{methodology}), when applied to these weights. Note that in these diagrams, a tensor is denoted by a circle, with each line emanating from the circle corresponding to a tensor dimension index. Also, connecting two index lines implies a summation over the connected indices. 

\section{Proposed Methodology}\label{methodology}
By exploiting the intuitions behind MHA \cite{clark2019does, vig2019multiscale,vig2019analyzing}, we propose a novel and intuitive multi-head tensorisation method. It first tensorises the MHA weights into a set of 3D tensors, each corresponding to an attention head. Then, Tucker decomposition is applied to the tensorised weights of each attention head in a single transformer layer, while sharing a common set of factor matrices across such decompositions. Uniquely, this ensures that the weights of the different attention heads are in a shared subspace characterised by a common set of factor matrices. By structurally denoising the attention weights using a shared higher-dimensional low-rank structure, our proposed framework is found to both improve LLM reasoning and simultaneously achieve compression in the MHA blocks of LLMs. 

\subsection{Multi-head Tensorisation}\label{tensorisation}
Despite their multi-head nature, the MHA weight matrices are usually stored in 2D formats to accelerate computation. To build a tensor from those matrices, we tensorise the model by first folding the weights into higher-dimensional formats, before applying tensor decompositions to compress the resulting weight tensor. To this end, we develop a multi-head tensorisation technique based on the intuitions about MHA, in order to naturally tensorise the original 2D query, key, value, and output projection weight matrices into a set of 3D tensors. More specifically, as shown in Step 1 and 2 in the left part of Fig. \ref{fig:methodology}, the tensorisation process starts by splitting the four global weight matrices in a single transformer layer, $\mathbf{W}^Q, \mathbf{W}^K, \mathbf{W}^V, \text{and } {\mathbf{W}^O}^T \in \mathbb{R}^{d_{model} \times h \cdot d_v}$, into the local sub-matrices, $\mathbf{W}^Q_i, \mathbf{W}^K_i, \mathbf{W}^V_i, {\mathbf{W}^O_i}^T \in \mathbb{R}^{d_{model} \times  d_v}$, belonging to each attention head $i$, where $1\leq i \leq h$. Next, for each attention head, the four 2D sub-matrices $\mathbf{W}^Q_i, \mathbf{W}^K_i, \mathbf{W}^V_i, {\mathbf{W}^O_i}^T$ are stacked into a 3D tensor, $\mathcal{W}_i \in \mathbb{R}^{d_{model} \times d_v \times 4}$, as
\begin{equation}
    \mathcal{W}_{i [:,:,j]} = 
    \begin{cases} 
        \mathbf{W}^Q_i & \text{if } j = 1, \\
        \mathbf{W}^K_i & \text{if } j = 2, \\
        \mathbf{W}^V_i & \text{if } j = 3, \\
        {\mathbf{W}^O_i}^T & \text{if } j = 4. \\
    \end{cases}
    \label{eq:single_head_weight}
\end{equation}
This process can then repeated for all $h$ heads before stacking all tensors $\{\mathcal{W}_i \}_{i=1}^h$ together in order to tensorise all MHA weight matrices of  a single transformer layer into a 4D tensor, $\mathcal{W}_{all} \in \mathbb{R}^{d_{model} \times d_v \times 4 \times h}$, as
\begin{equation}
    \mathcal{W}_{all[:,:,:,i]} = \mathcal{W}_i ,\text{ for  } 1\leq i \leq h.
    \label{eq:allweights}
\end{equation}
Such a multi-head tensorisation process converts the attention weight matrices into a higher-dimensional format to prepare for the utilisation of Tucker decomposition.

\subsection{Tucker Decomposition with Shared Factor Matrices}
As shown in Fig. \ref{fig:methodology}, Tucker decomposition decomposes a higher-dimensional tensor into a small-scale core tensor, which is of the same order as the original large-scale tensor, and multiple factor matrices. Physically, the core tensor represents the variability information in a subspace designated by the factor matrices. This makes it possible to apply Tucker decomposition to multiple attention weight tensors, to enable sharing the same set of factor matrices in order to enforce different information to reside within the same subspace.

To this end, in our approach, Tucker decomposition is applied to each of the $h$ 3D tensors, $\{\mathcal{W}_i \}_{i=1}^h$, defined in Eq. (\ref{eq:single_head_weight}), while sharing a common set of factor matrices, $\mathbf{U}^{(1)} \in \mathbb{R}^{d_{model} \times R_1}, \mathbf{U}^{(2)} \in \mathbb{R}^{d_{v} \times R_2}, \mathbf{U}^{(3)} \in \mathbb{R}^{4 \times R_3}$. With the variability information of the $i$-th attention head weights after Tucker decomposition being contained in a 3D tensor, $\mathcal{G}_i \in \mathbb{R}^{R_1 \times R_2 \times R_3}$, the weights of each attention head, $\mathcal{W}_i$, can be expressed as
\begin{equation}
    \mathcal{W}_i = \mathcal{G}_i \times_1 \mathbf{U}^{(1)} \times_2 \mathbf{U}^{(2)} \times_3 \mathbf{U}^{(3)}, \text{ for } 1\leq i \leq h.
\label{eq:tucker_with_shared_factors_1}
\end{equation}
Notice that this expression can be conveniently written as a special variant of the Tucker decomposition of a 4D tensor, in the form
\begin{equation}
    \mathcal{W}_{all} = \mathcal{G}_{all} \times_1 \mathbf{U}^{(1)} \times_2 \mathbf{U}^{(2)} \times_3 \mathbf{U}^{(3)} \times_4 \mathbf{I},
\label{eq:tucker_with_shared_factors}
\end{equation}
or in its element-wise definition form as 
\begin{equation}
\begin{split}
&\mathcal{W}_{all[i_1, i_2, i_3, i_4]} =  \\
&\quad \sum_{r_1=1}^{R_1} \sum_{r_2=1}^{R_2} \sum_{r_3=1}^{R_3} \mathcal{G}_{all [r_1, r_2, r_3, i_4]} \mathbf{U}^{(1)}_{[i_1, r_1]} \mathbf{U}^{(2)}_{[i_2, r_2]} \mathbf{U}^{(3)}_{[i_3, r_3]},
\end{split}
\label{eq:tucker_with_shared_factors_index_notation}
\end{equation}
where $\mathcal{W}_{all} \in \mathbb{R}^{d_{model} \times d_v \times 4 \times h}$ is defined in Eq. (\ref{eq:allweights}) and represents the tensor containing all attention weights in a single transformer layer, while $\mathbf{U}^{(1)}, \mathbf{U}^{(2)}, \text{and } \mathbf{U}^{(3)}$ are the shared  factor matrices. The term $\mathbf{I} \in \mathbb{R}^{h\times h}$ is an identity matrix which can be omitted, and the 4D core tensor, $\mathcal{G}_{all} \in \mathbb{R}^{R_1 \times R_2 \times R_3 \times h}$, is defined as 
\begin{equation}
    \mathcal{G}_{all [:,:,:,i]} = \mathcal{G}_i ,\text{ for  } 1\leq i \leq h.
    \label{eq:tucker_core_all}
\end{equation}
The denoising process is performed by approximating the original weight tensors of each attention head using the Tucker decomposition; in other words, given a set of multilinear ranks, we denoise the tensorised MHA weights by minimizing 
\begin{equation}
\frac{1}{2} \left\| \sum_{i=1}^h \mathcal{W}_i - \sum_{i=1}^h \mathcal{G}_i \times_1 \mathbf{U}^{(1)} \times_2 \mathbf{U}^{(2)} \times_3 \mathbf{U}^{(3)} \right\|^2_F
\label{eq:compact_tucker_objective_ori}
\end{equation}
or in an equivalent format
\begin{equation}
\frac{1}{2} \left\|\mathcal{W}_{all} - \mathcal{G}_{all} \times_1 \mathbf{U}^{(1)} \times_2 \mathbf{U}^{(2)} \times_3 \mathbf{U}^{(3)} \right\|^2_F.
\label{eq:compact_tucker_objective}
\end{equation}
In practice, we utilised the TensorLy library \cite{tensorly} to implement this special variant of Tucker decomposition based on the Higher Order Orthogonal Iterations (HOOI) algorithm \cite{de2000best}.
\begin{rem}
Observe from Eq. (\ref{eq:compact_tucker_objective_ori}) that the weights of each attention head are sharing a common set of factor matrices $\mathbf{U}^{(1)}, \mathbf{U}^{(2)}$ and $\mathbf{U}^{(3)}$. At the same time, each attention head is assigned its own Tucker core tensor. We conjecture that this design aligns with the intuition that attention heads within a single transformer layer capture patterns at similar abstraction levels with different specialisations.
\end{rem} 

\begin{rem}
The tensor decomposition process in Eqs. (\ref{eq:tucker_with_shared_factors}) - (\ref{eq:compact_tucker_objective}) allows us to structurally denoise the attention weight matrices according to a shared higher-dimensional low-rank structure. Additionally, this also allows for parameter compression, by representing the original weight tensor through smaller-sized factors.
\end{rem}


\section{Experiments}
Comprehensive experiments were conducted to verify the performance of our proposed framework on four benchmark reasoning datasets and three LLMs; this includes both the encoder-only and decoder-only architectures. In the experiments, our framework was applied in a layer-selective fashion similar to \cite{sharma2023truth}, to enable fair comparisons. For example, we applied the proposed method in Section \ref{methodology} to only a single transformer layer at a time. The results demonstrate that our model is capable of significantly improving the reasoning capabilities of the original LLM while enabling their parameter compression. Furthermore, experimental results show that our method can be used in conjunction with existing FFN-only compression methods, such as LASER \cite{sharma2023truth}, which improved the reasoning abilities of LLMs by denoising the FFN weights. Finally, we validated our framework with an ablation study which compares the proposed approach against that of denoising separate tensors for each query, key, value and output weight matrices in a transfomer layer. The experiments were conducted using NVIDIA A100 GPUs.

\begin{table}[htbp]
\centering
\caption{Performance comparison of RoBERTa, GPT-J, and LLaMA2, with and without applying our method, tested on four benchmark datasets. The highest accuracy $\uparrow$ and the lowest loss $\downarrow$ for each model and dataset combination are denoted in bold. CR $\uparrow$ stands for compression rate of the MHA parameters in a transformer layer. ``-" indicates no compression.}
\renewcommand{\arraystretch}{1.2}
\setlength{\tabcolsep}{2.5pt}
\begin{tabular}{|c c|p{0.95cm}<{\centering}|p{0.8cm}<{\centering}!{\vrule width 1pt}p{0.95cm}<{\centering}|p{0.8cm}<{\centering}!{\vrule width 1pt}p{0.95cm}<{\centering}|p{0.8cm}<{\centering}|}
\hline
& & \multicolumn{6}{c|}{Model Name} \\
\quad Dataset  & & \multicolumn{2}{c!{\vrule width 1pt}}{RoBERTa} & \multicolumn{2}{c!{\vrule width 1pt}}{GPT-J} & \multicolumn{2}{c|}{LLaMA2} \\ 
& &  Original & Ours & Original & Ours & Original & Ours \\ \hline
\multirow{3}{*}{\textit{HotPotQA}} & Acc & 6.1  & \textbf{7.33} & 19.6 & \textbf{20.15}  & 16.5  & \textbf{18.44}  \\ 
 & Loss & 10.99 & \textbf{10.00} & \textbf{3.40}  & 4.49 & \textbf{3.15} & 9.80 \\
 & CR & - & 1.12 & - & 247.30 & - & 3.54 \\ \hline
\multirow{3}{*}{\textit{FEVER}} & Acc & 50.0 & \textbf{50.45}  & 50.2  & \textbf{58.94} & 59.3 &  \textbf{66.75} \\ 
 & Loss & 2.5 & \textbf{1.47} & 1.24  & \textbf{1.02} & 1.02 & \textbf{1.01} \\ 
  & CR & - & 3.74 & - & 14.69 & - & 3.54 \\ \hline
 \multirow{3}{*}{\makecell{\textit{Bios} \\ \textit{Profession}}} & Acc & 64.5 & \textbf{72.57} & 75.6 & \textbf{81.18} & 85.0  & \textbf{86.61} \\ 
& Loss & \textbf{4.91} & 6.64 & 4.64  & \textbf{4.57} & \textbf{4.19}  & 4.54 \\ 
& CR & - & 8.78 & - & 74.68 & - & 3.54 \\ \hline
 \multirow{3}{*}{\makecell{\textit{BigBench-} \\ \textit{WikidataQA}}} & Acc & 28.0 & \textbf{32.72}  & 51.8 & \textbf{68.81} &  59.5 & \textbf{60.37}  \\ 
 & Loss & 9.07 & \textbf{8.72} & 3.52 & \textbf{2.63} & 4.19  & \textbf{2.38} \\ 
 & CR & - & 2.52  & - & 46.77 & - & 5.81 \\ \hline
\end{tabular}
\label{tab:performance}
\end{table}

\subsection{LLM Models}
We tested our proposed method on three LLM models: RoBERTa 125M\cite{liu2019RoBERTa}, GPT-J 6B \cite{gpt-j}, and LLaMA 2 7B \cite{llama2}.
\begin{itemize}
    \item RoBERTa has an encoder-only architecture and predicts missing tokens within a given context. Thus, we appended five $<$\texttt{mask}$>$ tokens to each question before inputting it to the RoBERTa model.
    \item Both GPT-J and LLaMA 2 are decoder-only models and can directly generate tokens autoregressively, by predicting the next token based on the preceding tokens.
\end{itemize}

\begin{table*}
\centering
\caption{Performance of stand-alone and hybrid methods for LLM compression and reasoning. Case 1: LASER applied on one weight matrix of the FFN block; Case 2: LASER applied on both the MHA and FFN blocks; Case 3: Our method applied on the MHA block; and LASER applied on the FFN block. The highest accuracy $\uparrow$ and the lowest loss $\downarrow$ of each model and dataset combination are denoted in bold.}
\begin{tabular}{|c c|c|c|c!{\vrule width 1pt}c|c|c!{\vrule width 1pt}c|c|c|}
\hline
& & \multicolumn{9}{c|}{Model Name} \\ 
\centering Dataset & & \multicolumn{3}{c!{\vrule width 1pt}}{RoBERTa} & \multicolumn{3}{c!{\vrule width 1pt}}{GPT-J} & \multicolumn{3}{c|}{LLaMA2} \\ 
& & Case 1 & Case 2  & Case 3 & Case 1 & Case 2 & Case 3 & Case 1 & Case 2 & Case 3 \\
& & & & (Ours) & & & (Ours) & & & (Ours) \\ \hline
\multirow{ 2}{*}{\textit{HotPotQA}} & Acc  & 6.7 & 5.24 & \textbf{7.05} & 19.5 & 19.62 & \textbf{19.91} & 17.2 & 18.88 & \textbf{19.22} \\ 
 & Loss & 10.53 & \textbf{8.60} & 9.87 & \textbf{3.39} & 5.08 & 5.07 & \textbf{2.97} & 9.33 & 9.99 \\ \hline
\multirow{ 2}{*}{\textit{FEVER}} & Acc  & 52.3 & 53.6 & \textbf{55.23} & 56.2 & 55.59 & \textbf{58.98} & 64.5 & 65.13  & \textbf{66.39} \\ 
 & Loss & 1.76 & \textbf{1.18} & 2.61 & \textbf{1.27} & 1.28 & 1.39 & \textbf{0.91} & 1.11 & 1.33 \\ \hline
\multirow{ 2}{*}{\textit{Bios Profession}} & Acc  & 72.5 & 71.14 & \textbf{72.51} & 82.1 & 81.28 & \textbf{82.52} & 86.7 & 86.07 & \textbf{87.07} \\ 
 & Loss & \textbf{6.44} & 6.62 & 7.42 & 4.91 & 4.61 & \textbf{4.52} & \textbf{4.05} & 4.20 & \textbf{4.05} \\ \hline
\multirow{ 2}{*}{\textit{BigBench-WikidataQA}} & Acc  & 30.7 & 34.49 & \textbf{37.40} & 65.9 & 65.68 & \textbf{68.20} & \textbf{62.0} & 61.21 & 61.78 \\ 
& Loss & \textbf{7.69} & 8.25 & 7.86 & 2.86 & 2.89  & \textbf{2.59} & \textbf{2.31} & 2.35 & 2.34 \\ \hline
\end{tabular}
\label{tab:exp2}
\end{table*}

\subsection{Datasets} \label{datasets}
We adopted the same data pre-processing processes as in LASER \cite{sharma2023truth} for the 4 benchmark reasoning datasets, \textit{HotPotQA} \cite{yang2018HotPotQA}, \textit{FEVER} \cite{thorne2018FEVER}, \textit{Bios Profession} \cite{de2019bias}, and \textit{BigBench-WikidataQA} \cite{bowman2015large}.
\paragraph{\textit{HotPotQA}} 
The \textit{HotPotQA} is a large-scale question-answering dataset. Question-answer pairs were extracted from its training and validation sets. The input texts were tokenized using the LLaMA 2 \cite{llama2} tokenizer with samples exceeding 15 tokens being discarded. Prompts were formatted as ``$<$\texttt{question}$>$ The answer is”, where $<$\texttt{question}$>$ was replaced by the actual question. The model was allowed to generate up to $\texttt{max\_len}$ tokens, and if the answer appearred in the generated text, it was considered correct.
\paragraph{\textit{FEVER}}
The Fact Extraction and VERification (\textit{FEVER}) dataset  was developed to evaluate fact-checking systems. It contains binary labels: 0 (false) and 1 (true). Claims with conflicting labels were filtered to ensure consistency. Question-answer pairs were extracted from its validation and test sets. Prompts were structured as: “Consider the following claim: $<$\texttt{claim}$>$. Is this claim true or false? The claim is”, where $<$\texttt{claim}$>$ was replaced by the actual statement. If the probability of a claim being true exceeded that of it being false, the claim was classified as true and otherwise false.
\paragraph{\textit{Bios Profession}}
The \textit{Bios Profession} dataset is a benchmark for analyzing gender biases in occupational and professional contexts. The task involves predicting the profession given a biography. Question-answer pairs were derived from the validation set, focusing on the same 10 professions used in the LASER paper. Prompts were structured as: ``Consider the following text: $<$\texttt{bio}$>$. What is the profession of the person in this text? The profession of this person is" where $<$\texttt{bio}$>$ was replaced by the actual biography. The model was tasked to output the profession with the highest probability among the 10 possible professions.
\paragraph{\textit{BigBench-WikidataQA}}
In Beyond the Imitation Game Benchmark (BIG-Bench) dataset, the WikidataQA subset focuses on question answering (QA) using structured knowledge from Wikidata. Question-answer pairs were derived from the train and validation set. The model was allowed to generate up to $\texttt{max\_len}$ tokens, and if the answer appearred in the generated text, it was considered correct.

\subsection{Experimental Results}
To assess the enhancement in the reasoning capabilities of LLMs with our proposed framework, we applied it to three LLMs and evaluated their performance on the four reasoning datasets mentioned in Section \ref{datasets}. Table \ref{tab:performance} shows the accuracy, loss, and compression ratio of the LLMs with and without our method applied to the MHA block. Accuracy is the key evaluation metric for model reasoning performances and is measured by the percentage of correctly predicted instances in the test set. Test accuracy was evaluated using the last 20\% of data in each dataset. Loss is an indirect performance measure and represents the uncertainty of the model, i.e., the deviation from the true target distribution. The loss is included for completeness and measured by the negative log-likelihood of the ground truth token. The Compression Ratio (CR) quantifies the reduction in model size and is computed as $\text{CR}= \frac{N_{original}}{N_{compressed}}$, where $N_{original}$ and $N_{compressed}$ represent the total number of parameters of the MHA weights in an original single transformer layer and a single compressed transformer layer, respectively. Table \ref{tab:performance} shows that our proposed method consistently improved the performance of the three original models on all four reasoning datasets, in terms of test accuracy, and achieved compression rates of the MHA weights up to $247.3$ times. \par 

\begin{rem}
Our method is applied to the MHA blocks so that it can be used in conjunction with existing methods  to further improve the reasoning capabilities of LLMs. For example, in combination with LASER \cite{sharma2023truth} which achieves best performance when applied to the FFN blocks.
\label{rem:hypbrid}
\end{rem}

To evaluate such ``hybrid" scenarios as mentioned in Remark \ref{rem:hypbrid}, we investigated the following three cases: 
\begin{itemize}
    \item Case 1: LASER\cite{sharma2023truth} was applied to one matrix in the FFN block;
    \item Case 2: LASER\cite{sharma2023truth} was applied to all matrices in the FFN and MHA blocks;
    \item Case 3: Our method was applied to the MHA block; LASER\cite{sharma2023truth} was applied to matrices in the FFN block.
\end{itemize}
The results for Case 1 are directly quoted from the best performances reported in \cite{sharma2023truth}. Furthermore, as pointed out by the authors in \cite{sharma2023truth}, applying their method multiple times to many weight matrices can yield further improvements. To this end, in Case 2, we applied the LASER method to both the FFN block and the MHA block to obtain the best performance of LASER and allow for a fair comparison with Case 3, where our method was applied to the MHA block while LASER was applied to the FFN block. Table \ref{tab:exp2} presents the results obtained for the above three cases. Case 3 achieves the highest test accuracies across the three models and four datasets, except with LLaMA2 on the \textit{BigBench-WikidataQA} dataset. This further highlights the intuition behind our approach which focuses on the MHA weights. Through a careful design of the tensorisation process and the utilisation of current domain knowledge about MHA, our method was capable of structurally denoising the MHA weights according to the intuition of having a shared high-dimensional subspace among the attention heads in a single transformer layer. This not only enhances the interpretability of our approach, but also demonstrates that our framework can be used as a versatile module, in conjunction with methods designed for the FFN weights only, in order to achieve enhanced reasoning and compression in LLMs.

\subsection{Ablation study}
To evaluate the impact of stacking the query, key, value, and output weight matrices into a tensor, we compared our framework to the scenario where the same methodology was applied to one of the four types of weight matrices in MHA -- query, key, value, and output. Table \ref{tab:ablation} shows that our proposed method was able to consistently achieve the best performance across all four reasoning datasets. This validates our underpinning conjecture of tensorising together all MHA weights in a transformer layer.

\begin{table}[t]
\centering
\caption{The impact of compressing the $\mathbf{W}^Q$, $\mathbf{W}^K$, $\mathbf{W}^V$, and $\mathbf{W}^O$ separately and together. Our proposed method compresses simultaneously all MHA weights in a transformer layer. The highest accuracy $\uparrow$ and the lowest loss $\downarrow$ of each dataset are denoted in bold.}
\renewcommand{\arraystretch}{1.2}
\setlength{\tabcolsep}{3pt}
\begin{tabular}{|c c|p{0.95cm}<{\centering}|p{0.8cm}<{\centering}|p{0.8cm}<{\centering}|p{0.8cm}<{\centering}|p{0.8cm}<{\centering}|p{0.8cm}<{\centering}|}
\hline
& & \multicolumn{6}{c|}{GPT-J} \\
\centering Dataset  & & Original & $\mathbf{W}^Q$ & $\mathbf{W}^K$  & $\mathbf{W}^V$ & $\mathbf{W}^O$ & Ours \\ \hline
\multirow{2}{*}{\textit{HotPotQA}} & Acc  & 19.6 &  19.19 & 19.25 & 19.70 & 19.62 & \textbf{20.15} \\ 
 & Loss & \textbf{3.4} & 4.45 & 4.45 & 4.43 & 4.44 & 4.49 \\ \hline
\multirow{2}{*}{\textit{FEVER}} & Acc & 50.2 & 54.41 & 53.40 & 55.86 & 56.07 & \textbf{58.94} \\ 
 & Loss & 1.24 & 1.22 & 1.22 & 1.23 & 1.15 & \textbf{1.02} \\ \hline
\multirow{2}{*}{\makecell{\textit{Bios} \\ \textit{Profession}}} & Acc & 75.6 & 76.06 & 74.97 & 79.39 & 79.71 & \textbf{81.18} \\ 
 & Loss & 4.64 & 4.54 & 4.59 & 4.46 & \textbf{4.41} & 4.57 \\ \hline
\multirow{2}{*}{\makecell{\textit{BigBench-} \\ \textit{WikidataQA}}} & Acc & 51.8 & 49.72 & 51.01 & 48.82 & 48.87 & \textbf{68.81} \\ 
 & Loss & 3.52 & 3.66 & 3.58 & 3.69 & 3.69 & \textbf{2.63} \\ \hline
\end{tabular}
\label{tab:ablation}
\end{table}

\section{Conclusion}
We have proposed a novel framework for the enhancement of the reasoning abilities of LLMs while simultaneously preforming parameter compression. This has been achieved by exploiting the domain knowledge and empirical evidence about multi-head attention in LLMs, along with a unique multi-head tensorisation and a special variant of the Tucker decomposition. In this way, our framework has explored structurally denoising the weights of each attention head in a transformer layer, according to a shared higher-dimensional low-rank structure among the attention heads. Consequently, this has ensured that the weights of each attention head encode different information within a common higher-dimensional subspace characterised by the common Tucker factor matrices. Our method has been shown to enable parameter compression and enhanced reasoning in both encoder-only and decoder-only LLMs, of which the parameter complexity ranges from hundreds of millions to billions of parameters. Additionally, our framework can be used in conjunction with existing methods that improve LLM reasoning, such as those based on denoising FFN weights, to yield further performance gains. Through an ablation study, we have validated the advantage and performance gain obtained by our proposed multi-head tensorisation process.
\paragraph{Limitations}
Similar to other existing weight denoising methods, we have found that for different datasets, our method achieves the best results under different hyperparameter settings. Our future work will focus on finding unified and generalisable hyperparameters settings for both our proposed method and other existing methods.

\bibliography{reference}
\bibliographystyle{ieeetr}

\end{document}